\documentclass[11pt]{article} 

\usepackage{url}
\usepackage{times}
\usepackage[square,sort,comma,numbers]{natbib}

\usepackage[utf8]{inputenc} 
\usepackage[T1]{fontenc}    
\usepackage{hyperref}       
\usepackage{url}            
\usepackage{booktabs}       
\usepackage{nicefrac}       
\usepackage{microtype}      
\usepackage{color}

\usepackage{graphicx} 

\begin{document}

\begin{center}
\large
{Stop Explaining Black Box Machine Learning Models for High Stakes Decisions and Use Interpretable Models Instead
\\} 
\normalsize
\vspace*{5pt}
\textsl{Cynthia Rudin \\ Duke University\\
cynthia@cs.duke.edu}
\end{center}

\begin{abstract}
Black box machine learning models are currently being used for high stakes decision-making throughout society, causing problems throughout healthcare, criminal justice, and in other domains. People have hoped that creating methods for explaining these black box models will alleviate some of these problems, but trying to \textit{explain} black box models, rather than creating models that are \textit{interpretable} in the first place, is likely to perpetuate bad practices and can potentially cause catastrophic harm to society. There is a way forward -- it is to design models that are inherently interpretable. This manuscript clarifies the chasm between explaining black boxes and using inherently interpretable models, outlines several key reasons why explainable black boxes should be avoided in high-stakes decisions, identifies challenges to interpretable machine learning, and provides several example applications where interpretable models could potentially replace black box models in criminal justice, healthcare, and computer vision.
\end{abstract}

\section{Introduction}

There has been an increasing trend in healthcare and criminal justice to leverage machine learning (ML) for high-stakes prediction applications that deeply impact human lives. Many of the ML models are black boxes that do not explain their predictions in a way that humans can understand. The lack of transparency and accountability of predictive models can have (and has already had) severe consequences; there have been cases of people incorrectly denied parole \citep{nyt-computers-crim-justice}, poor bail decisions leading to the release of dangerous criminals, ML-based pollution models stating that highly polluted air was safe to breathe \citep{McGough2018}, and generally poor use of limited valuable resources in criminal justice, medicine, energy reliability, finance, and in other domains \citep{VarshneyAl2016}. 

Rather than trying to create models that are inherently interpretable, there has been a recent explosion of work on ``Explainable ML,'' where a second (posthoc) model is created to explain the first black box model. This is problematic. Explanations are often not reliable, and can be misleading, as we discuss below. If we instead use models that are inherently interpretable, they provide their own explanations, which are faithful to what the model actually computes.

In what follows, we discuss the problems with Explainable ML, followed by the challenges in Interpretable ML. 
This document is mainly relevant to high-stakes decision making and troubleshooting models, which are the main two reasons one might require an interpretable or explainable model. Interpretability is a domain-specific notion \citep{Freitas:2014ic,kodratoff1994comprehensibility,huysmans:2011gq,ruping2006learning}, so there cannot be an all-purpose definition. Usually, however, an interpretable machine learning model is \textit{constrained in model form} so that it is either useful to someone, or obeys structural knowledge of the domain, such as monotonicity \citep[e.g.,][]{gupta2016monotonic}, causality, structural (generative) constraints, additivity \citep{LouCaGeHo13}, or physical constraints that come from domain knowledge. Interpretable models could use case-based reasoning for complex domains. Often for structured data, sparsity is a useful measure of interpretability, since humans can handle at most 7$\pm$2 cognitive entities at once \citep{miller1956magical,cowan2010magical}. Sparse models allow a view of how variables interact \textit{jointly} rather than individually. We will discuss several forms of interpretable machine learning models for different applications below, but there can never be a single definition; e.g., in some domains, sparsity is useful, and in others is it not. There is a spectrum between fully transparent models (where we understand how all the variables are \textit{jointly} related to each other) and models that are lightly constrained in model form (such as models that are forced to increase as one of the variables increases, or models that, all else being equal, prefer variables that domain experts have identified as important, see \citep{Wiens2018}).

A preliminary version of this manuscript appeared at a workshop, entitled ``Please Stop Explaining Black Box Machine Learning Models for High Stakes Decisions'' \citep{Rudin18}. 

\section{Key Issues with Explainable ML}
A black box model could be either (i) a function that is too complicated for any human to comprehend, or (ii) a function that is proprietary (see Appendix \ref{twotypes}). 
Deep learning models, for instance, tend to be black boxes of the first kind because they are highly recursive. 
As the term is presently used in its most common form, an explanation is a separate model that is supposed to replicate most of the behavior of a black box (e.g., ``the black box says that people who have been delinquent on current credit are more likely to default on a new loan''). Note that the term ``explanation'' here refers to an understanding of how a model works, as opposed to an explanation of how the world works. The terminology ``explanation'' will be discussed later; it is misleading.  

I am concerned that the field of interpretability/explainability/comprehensibility/transparency in machine learning has strayed away from the needs of real problems. This field dates back to the early 90's at least \citep[see][]{Freitas:2014ic,Holte93}, and there are a huge number of papers on interpretable ML in various fields (that often do not have the word ``interpretable'' or ``explainable'' in the title, as the recent papers do). 
Recent work on explainability of black boxes -- rather than interpretability of models -- contains and perpetuates critical misconceptions that have generally gone unnoticed, but that can have a lasting negative impact on the widespread use of machine learning models in society. Let us spend some time discussing this before discussing possible solutions. \\

\noindent\textbf{(i) It is a myth that there is necessarily a trade-off between accuracy and interpretability.}

There is a widespread belief that more complex models are more accurate, meaning that a complicated black box is necessary for top predictive performance. However, this is often not true, particularly when the data are structured, with a good representation in terms of naturally meaningful features. When considering problems that have structured data with meaningful features, there is often no significant difference in performance between more complex classifiers (deep neural networks, boosted decision trees, random forests) and much simpler classifiers (logistic regression, decision lists) after preprocessing. (Appendix \ref{AppendixSame} discusses this further.)
In data science problems, where structured data with meaningful features are constructed as part of the data science process, there tends to be little difference between algorithms, assuming that the data scientist follows a standard process for knowledge discovery \citep[such as KDD, CRISP-DM, or BigData, see][]{Fayyad96fromdata, crispdm, bigdata}. 

Even for applications such as computer vision, where deep learning has major performance gains, and where interpretability is much more difficult to define, some forms of interpretability can be imbued directly into the models without losing accuracy. This will be discussed more later in the Challenges section. Uninterpretable algorithms can still be useful in high-stakes decisions as part of the knowledge discovery process, for instance, to obtain baseline levels of performance, but they are not generally the final goal of knowledge discovery.

\begin{figure}
\centering
\includegraphics[width = 0.3\textwidth]{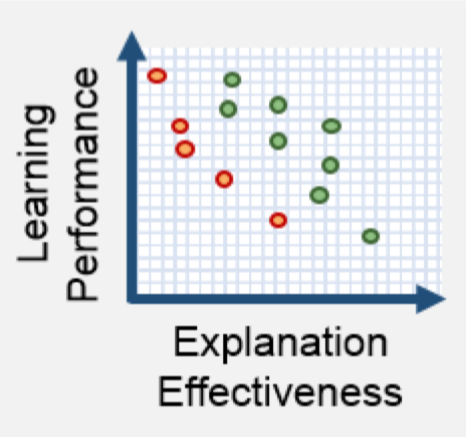}
\caption{A fictional depiction of the ``accuracy-interpretability trade-off,'' taken from the DARPA XAI (Explainable Artificial Intelligence) Broad Agency Announcement \citep{XAIBAA}.\label{fig:XAI}}	
\end{figure}

Figure \ref{fig:XAI}, taken from the DARPA Explainable Artificial Intelligence program's Broad Agency Announcement \citep{XAIBAA}, exemplifies a blind belief in the myth of the accuracy-interpretability trade-off. This not a ``real'' figure, in that it was not generated by any data. The axes have no quantification (there is no specific meaning to the horizontal or vertical axes). The image appears to illustrate an experiment with a static dataset, where several machine learning algorithms are applied to the same dataset. However, this kind of smooth accuracy/interpretability/explainability trade-off is atypical in data science applications with meaningful features. Even if one were to quantify the interpretability/explainability axis and aim to show that such a trade-off did exist, it is not clear what algorithms would be applied to produce this figure. (Would one actually claim it is fair to compare the 1984 decision tree algorithm CART to a 2018 deep learning model and conclude that interpretable models are not as accurate?) One can always create an artificial trade-off between accuracy and interpretability/explainability by removing parts of a more complex model to reduce accuracy, but this is not representative of the analysis one would perform on a real problem. It is also not clear why the comparison should be performed on a static dataset, because any formal process for defining knowledge from data \citep{Fayyad96fromdata, crispdm, bigdata} would require an iterative process, where one refines the data processing after interpreting the results. Generally, in the practice of data science, the small difference in performance between machine learning algorithms can be overwhelmed by the ability to interpret results and process the data better at the next iteration \citep{Hand}. In those cases, the accuracy/interpretability tradeoff is reversed -- more interpretability leads to better overall accuracy, not worse.

Efforts working within a knowledge discovery process led me to work in interpretable machine learning \citep{RudinETAL2010}. Specifically, I participated in a large-scale effort to predict electrical grid failures across New York City. The data were messy, including free text documents (trouble tickets), accounting data about electrical cables from as far back as the 1890's, inspections data from a brand new manhole inspections program; even the structured data were not easily integrated into a database, and there were confounding issues and other problems. Algorithms on a static dataset were at most 1\% different in performance, but the ability to interpret and reprocess the data led to significant improvements in performance, including correcting problems with the dataset, and revealing false assumptions about the data generation process. The most accurate predictors we found were sparse models with meaningful features that were constructed through the iterative process.

The belief that there is always a trade-off between accuracy and interpretability has led many researchers to forgo the \textit{attempt} to produce an interpretable model. This problem is compounded by the fact that researchers are now trained in deep learning, but not in interpretable machine learning. Worse, toolkits of machine learning algorithms offer little in the way of useful interfaces for interpretable machine learning methods.


To our knowledge, all recent review and commentary articles on this topic imply (implicitly or explicitly) that the trade-off between interpretability and accuracy generally occurs. It could be possible that there are application domains where a complete black box is required for a high stakes decision. As of yet, I have not encountered such an application, despite having worked on numerous applications in healthcare and criminal justice \citep[e.g.,][]{RudinUs18}, energy reliability \citep[e.g.,][]{RudinETAL2010}, and financial risk assessment \citep[e.g.,][]{ChenEtAlFICO2018}. \\

\noindent \textbf{(ii) Explainable ML methods provide explanations that are not faithful to what the original model computes.}

Explanations must be wrong. They cannot have perfect fidelity with respect to the original model. If the explanation was completely faithful to what the original model computes, the explanation would equal the original model, and one would not need the original model in the first place, only the explanation. (In other words, this is a case where the original model would be interpretable.) 
This leads to the danger that any explanation method for a black box model can be an inaccurate representation of the original model in parts of the feature space. \citep[See also for instance,][among others.]{MittelstadtEtAl19}

 An inaccurate (low-fidelity) explanation model limits trust in the explanation, and by extension, trust in the black box that it is trying to explain. An explainable model that has a 90\% agreement with the original model indeed explains the original model most of the time. However, an explanation model that is correct 90\% of the time is wrong 10\% of the time. If a tenth of the explanations are incorrect, one cannot trust the explanations, and thus one cannot trust the original black box. 
 If we cannot know for certain whether our explanation is correct, we cannot know whether to trust either the explanation or the original model.

A more important misconception about explanations stems from the terminology ``explanation,'' which is often used in a misleading way, because explanation models do not always attempt to mimic the calculations made by the original model. Even an explanation model that performs almost identically to a black box model might use completely different features, and is thus not faithful to the computation of the black box. Consider a black box model for criminal recidivism prediction, where the goal is to predict whether someone will be arrested within a certain time after being released from jail/prison. Most recidivism prediction models depend explicitly on age and criminal history, but do not explicitly depend on race. Since criminal history and age are correlated with race in all of our datasets, a fairly accurate explanation model could construct a rule such as ``This person is predicted to be arrested because they are black.'' This might be an accurate explanation model since it correctly mimics the predictions of the original model, but it would not be faithful to what the original model computes. This is possibly the main flaw identified by criminologists \citep{Flores16} in the ProPublica analysis \citep{propublica2016,LarsonMaKiAn16} that accused the proprietary COMPAS recidivism model of being racially biased. COMPAS (Correctional Offender Management Profiling for Alternative Sanctions) is a proprietary model that is used widely in the U.S. Justice system for parole and bail decisions. ProPublica created a linear explanation model for COMPAS that depended on race, and then accused the black box COMPAS model of depending  on race, conditioned on age and criminal history. In fact, COMPAS seems to be nonlinear, and it is entirely possible that COMPAS does not depend on race (beyond its correlations with age and criminal history) \citep{RudinWaCo18}. ProPublica's linear model was not truly an ``explanation'' for COMPAS, and they should not have concluded that their explanation model uses the same important features as the black box it was approximating. (There will be a lot more discussion about COMPAS later in this document.)

An easy fix to this problem is to change terminology. Let us stop calling approximations to black box model predictions \textit{explanations}. For a model that does not use race explicitly, an automated explanation ``This model predicts you will be arrested because you are black'' is not an explanation of what the model is actually doing, and would be confusing to a judge, lawyer or defendant. Recidivism prediction will be discussed more later, as it is a key application where interpretable machine learning is necessary. In any case, it can be much easier to detect and debate possible bias or unfairness with an interpretable model than with a black box. Similarly, it could be easier to detect and avoid data privacy issues with interpretable models than black boxes.
Just as in the recidivism example above, many of the methods that claim to produce \textit{explanations} instead compute useful \textit{summary statistics of predictions} made by the original model. Rather than producing explanations that are faithful to the original model, they show trends in how predictions are related to the features. Calling these ``summaries of predictions,'' ``summary statistics,'' or ``trends'' rather than ``explanations'' would be less misleading.\\


\noindent \textbf{(iii) Explanations often do not make sense, or do not provide enough detail to understand what the black box is doing.}

\begin{figure}
\centering
\includegraphics[width = 0.8\textwidth]{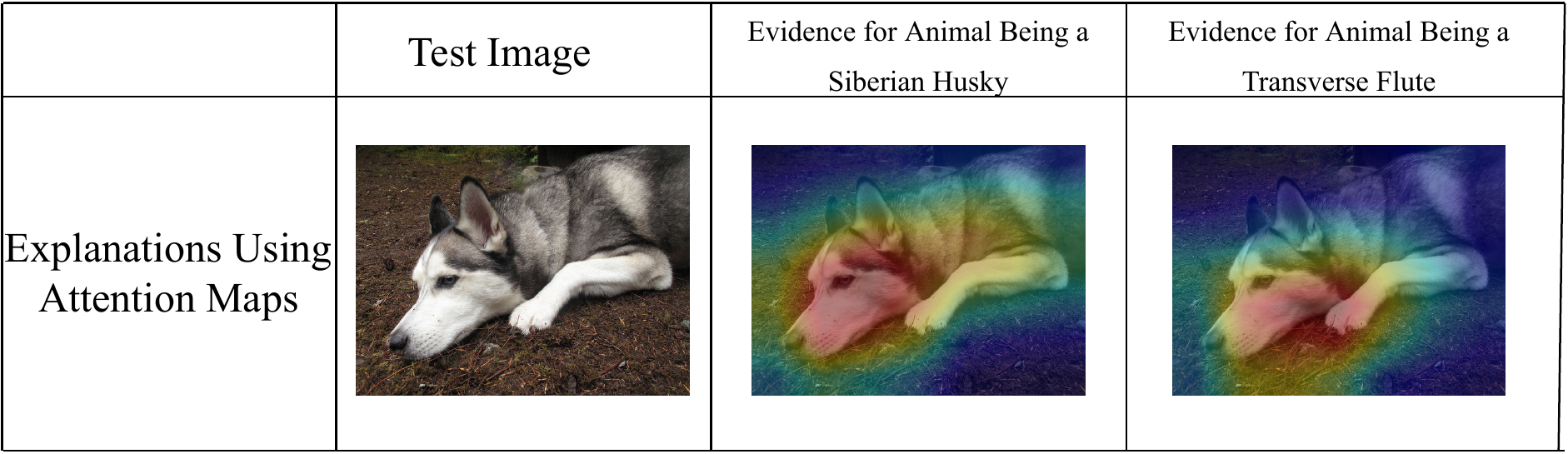}
\caption{Saliency does not explain anything except where the network is looking. We have no idea why this image is labeled as either a dog or a musical instrument when considering only saliency. The explanations look essentially the same for both classes. Figure credit: Chaofan Chen and \cite{Checkermallow}. \label{fig:saliency}}	
\end{figure}

Even if both models are correct (the original black box is correct in its prediction and the explanation model is correct in its approximation of the black box's prediction), it is possible that the explanation leaves out so much information that it makes no sense. 
I will give an example from image processing, for a low-stakes decision (not a high-stakes decision where explanations are needed, but where explanation methods are often demonstrated). 
Saliency maps are often considered to be explanatory. Saliency maps can be useful to determine what part of the image is being omitted by the classifier, but this leaves out all information about how relevant information \textit{is} being used. Knowing where the network is looking within the image does not tell the user what it is doing with that part of the image, as illustrated in Figure \ref{fig:saliency}. In fact, the saliency maps for multiple classes could be essentially the same; in that case, the explanation for why the image might contain a Siberian husky would be the same as the explanation for why the image might contain a transverse flute.

An unfortunate trend in recent work is to show explanations only for the observation's \textit{correct} label when demonstrating the method (e.g., Figure \ref{fig:saliency} would not appear). Demonstrating a method using explanations only for the correct class is misleading. This practice can instill a false sense of confidence in the explanation method and in the black box. Consider, for instance, a case where the explanations for multiple (or all) of the classes are identical. 
This situation would happen often when saliency maps are the explanations, because they tend to highlight edges, and thus provide similar explanations for each class. These explanations could be identical even if the model is \textit{always} wrong. Then, showing only the explanations for the image's correct class misleads the user into thinking that the explanation is useful, \textit{and} that the black box is useful, even if neither one of them are. 

Saliency maps are only one example of  explanations that are so incomplete that they might not convey why the black box predicted what it did. Similar arguments can be made with other kinds of explanation methods. Poor explanations can make it very hard to troubleshoot a black box. 
\\

\noindent \textbf{(iv) Black box models are often not compatible with situations where information outside the database needs to be combined with a risk assessment.}

In high stakes decisions, there are often considerations outside the database that need to be combined with a risk calculation. 
For instance, what if the circumstances of the crime are much worse than a generic assigned charge? There are often circumstances whose knowledge could either increase or decrease someone's risk. But if the model is a black box, it is very difficult to manually calibrate how much this additional information should raise or lower the estimated risk. This issue arises constantly; for instance, the proprietary COMPAS model used in the U.S. Justice System for recidivism risk prediction does not depend on the seriousness of the current crime \citep{northpointe,RudinWaCo18}. Instead, the judge is instructed to somehow manually combine current crime with COMPAS. Actually, it is possible that many judges do not know this fact. If the model were transparent, the judge could see directly that the seriousness of the current crime is not being considered in the risk assessment.\\


\noindent \textbf{(v) Black box models with explanations can lead to an overly complicated decision pathway that is ripe for human error.}

Typographical errors seem to be common in computing COMPAS, and these typographical errors sometimes determine bail decision outcomes \citep{nyt-computers-crim-justice,RudinWaCo18}. This exemplifies an important drawback of using overly complicated black box models for recidivism prediction -- they may be incorrectly calculated in practice. The computation of COMPAS requires 130+ factors. If typographical errors by humans entering these data into a survey occur at a rate of 1\%, then more than 1 out of every 2 surveys on average will have at least one typographical error. The multitude of typographical errors has been argued to be a type of \textit{procedural unfairness}, whereby two individuals who are identical might be randomly given different parole or bail decisions. These types of errors have the potential to reduce the in-practice accuracy of these complicated models.  


On the separate topic of model troubleshooting, an overly complicated black box model may be flawed but we do not know it, because it is difficult to troubleshoot. Having an (incomplete) explanation of it may not help, and now we must troubleshoot two models rather than one (the black box model and the explanation model).

In the next section, we completely switch gears. We will discuss reasons why so many people appear to advocate for black box models with separate explanation models, rather than inherently interpretable models -- even for high-stakes decisions. 

\section{Key Issues with Interpretable ML}

There are many cases where black boxes with explanations are preferred over interpretable models, even for high-stakes decisions. However, for most applications, I am hopeful that there are ways around some of these problems, whether they are computational problems, or problems with training of researchers and availability of code. The first problem, however, is currently a major obstacle that I see no way of avoiding other than through policy, as discussed in the next section.\\

\noindent \textbf{(i) Corporations can make profits from the intellectual property afforded to a black box.} 

Companies that charge for individual predictions could find their profits obliterated if an interpretable model were used instead. 

Consider the COMPAS proprietary recidivism risk prediction tool discussed above that is in widespread use in the U.S. Justice System for predicting the probability that someone will be arrested after their release \citep{northpointe}. 

The COMPAS model is equally accurate for recidivism prediction as the very simple three rule interpretable machine learning model involving only age and number of past crimes shown in Figure \ref{fig:CORELSModel} below. However, there is no clear business model that would suggest profiting from the simple transparent model. The simple model in Figure \ref{fig:CORELSModel} was created from an algorithm called Certifiably Optimal Rule Lists (CORELS) that looks for if-then patterns in data. Even though the model in Figure \ref{fig:CORELSModel} looks like a rule of thumb that a human may have designed without data, it is instead a full-blown machine learning model. A qualitative comparison of the COMPAS and CORELS models is in Table \ref{table:comparisontable}. Standard machine learning tools and interpretable machine learning tools seem to be approximately equally accurate for predicting recidivism, even if we define recidivism in many different ways, for many different crime types \citep{ZengUsRu2017, tollenaar2013method}. This evidence, however, has not changed the momentum of the justice system towards proprietary models. As of this writing, California has recently eliminated its cash bail system, instead enforcing that decisions be made by algorithms; it is unclear whether COMPAS will be the algorithm used for this, despite the fact that it is not known to be any more accurate than other models, such as the simple CORELS model in Figure \ref{fig:CORELSModel}.

\begin{figure}[h]
\hrule
\begin{eqnarray*}
\textrm{IF} & \textrm{age between 18-20 and sex is male}        &              \textrm{THEN predict arrest (within 2 years)}\\
\textrm{ELSE IF} & \textrm{age between 21-23 and 2-3 prior offenses} & \textrm{THEN predict arrest}\\
\textrm{ELSE IF} & \textrm{more than three priors}                           &     \textrm{THEN predict arrest}\\
\textrm{ELSE} 					      &   	          \textrm{predict no arrest.}
\end{eqnarray*}
\hrule
\caption{This is a machine learning model from the Certifiably Optimal Rule Lists (CORELS) algorithm \citep{angelino2018}. This model is the minimizer of a special case of Equation \ref{eq:optim} discussed later in the challenges section. CORELS' code is open source and publicly available at http://corels.eecs.harvard.edu/, along with the data from Florida needed to produce this model. \label{fig:CORELSModel}}
\end{figure}

\begin{table*}[h]
\centering
\begin{tabular}{|c|c|}
\hline
COMPAS &	CORELS\\
\hline
black box &	full model is in Figure \ref{fig:CORELSModel}\\
130+ factors	& only age, priors, (optional) gender\\
might include socio-economic info & 	no other information\\
expensive (software license), & free, transparent\\
within software used in U.S. Justice System	 &\\
\hline
\end{tabular}
\caption{Comparison \label{table:comparisontable} of COMPAS and CORELS models. Both models have similar true and false positive rates and true and false negative rates on data from Broward County, Florida.}
\end{table*}

COMPAS is not a machine learning model -- it was not created by any standard machine learning algorithm. It was designed by experts based on carefully designed surveys and expertise, and it does not seem to depend heavily on past criminal history \citep{RudinWaCo18}. Interestingly, if the COMPAS model were not proprietary, its documentation \citep{northpointe} indicates that it would actually be an interpretable predictive model. (It is a black box of the second type -- proprietary -- but not the first type -- complicated -- discussed above.) Revealing this model, however, would be revealing a trade secret.

Let us switch examples to consider the proprietary machine learning model by BreezoMeter, used by Google during the California wildfires of 2018, which predicted air quality as ``good -- ideal air quality for outdoor activities,'' when air quality was dangerously bad according to multiple other models \citep{McGough2018}, and people reported their cars covered in ash. The Environmental Protection Agency's free, vigorously-tested air quality index would have provided a reliable result \citep{Mannshardt2018}. How could BreezoMeter's machine learning method be so badly wrong and put so many in danger? We will never find out, but BreezoMeter, who has probably made a profit from making these predictions, may not have developed this new technology if its models were forced to be transparent.

In medicine, there is a trend towards blind acceptance of black box models, which will open the door for companies to sell more models to hospitals. For instance, radiology and in-hospital patient monitoring are areas of medicine that stand to gain tremendously by automation; humans cannot process data fast enough or rapidly enough to compete with machines. However, in trusting these automated systems, we must also trust the full database on which they were trained, the processing of the data, along with the completeness of the database. If the database does not represent the full set of possible situations that can arise, then the model could be making predictions in cases that are very different from anything it was trained on.
An example of where this can go wrong is given by Zech et al. \citep{Zech2018}, who noticed that their neural network was picking up on the word ``portable'' within an x-ray image, representing the type of x-ray equipment rather than the medical content of the image. If they had used an interpretable model, or even an explainable model, this issue would never have gone unnoticed. Zech et al. \citep{Zech2018} pointed out the issue of confounding generally; in fact, the plague of confounding haunts a vast number of datasets, and particularly medical datasets. This means that proprietary models for medicine can have serious errors. These models can also be fragile, in that if the model is used in practice in a slightly different setting than how it was trained (e.g., new x-ray equipment), accuracy can substantially drop.  

The examples of COMPAS, Breezometer, and black box medical diagnosis all illustrate a problem with the business model for machine learning. In particular, there is a conflict of responsibility in the use of black box models for high-stakes decisions: \textit{the companies that profit from these models are not necessarily responsible for the quality of individual predictions}. A prisoner serving an excessively long sentence due to a mistake entered in an overly-complicated risk score could suffer for years, whereas the company that constructed this complicated model is unaffected. On the contrary, the fact that the model was complicated and proprietary allowed the company to profit from it. In that sense, the model's designers are not incentivized to be careful in its design, performance, and ease of use. These are some of the same types of problems affecting the credit rating agencies who priced mortgages in 2008; that is, these are the same problems that contributed to the financial crisis in the United States at that time. 

One argument favoring black boxes is that keeping these models hidden prevents them from being gamed or reverse-engineered. It is not clear that this argument generally makes sense. In fact, the reason a system may be gamed is because it most likely was not designed properly in the first place, leading to a form of Goodhart's law if it were revealed. Quoting from Chang et al. \citep{ChangEtAl2012} about product rating systems: ``If the ratings are accurate measures of quality, then making the ratings more transparent could have a uniformly positive impact: it would help companies to make better rated products, it would help consumers to have these higher quality products, and it would encourage rating companies to receive feedback as to whether their rating systems fairly represent quality.'' Thus, transparency could help improve the quality of the system, whereby attempting to game it would genuinely align with the overall goal of improvement. For instance, improving one's credit score should actually correspond to an improvement in creditworthiness.

Another argument favoring black boxes is the belief that ``counterfactual explanations'' of black boxes are sufficient. A counterfactual explanation describes a minimal change to the input that would result in the opposite  prediction. For instance, a possible counterfactual explanation might be ``your loan application was denied, but if you had \$1000 less debt, you would have qualified for the loan.'' This type of explanation can suffer from key issue (iv) discussed above, about combining information outside the database with the black box. In particular, the ``minimal'' change to the input might be different for different individuals. Appendix \ref{appendix:counterfactual} discusses in more depth why counterfactual explanations generally do not suffice for high stakes decisions of black boxes.\\



 \noindent \textbf{(ii) Interpretable models can entail significant effort to construct, in terms of both computation and domain expertise.}

As discussed above, interpretability usually translates in practice to a set of application-specific constraints on the model. Solving constrained problems is generally harder than solving unconstrained problems. Domain expertise is needed to construct the definition of interpretability for the domain, and the features for machine learning. For data that are unconfounded, complete, and clean, it is much easier to use a black box machine learning method than to troubleshoot and solve computationally hard problems. However, for high-stakes decisions, analyst time and computational time are less expensive than the cost of having a flawed or overly complicated model. That is, it is worthwhile to devote extra effort and cost into constructing a high-quality model. But even so, many organizations do not have analysts who have the training or expertise to construct interpretable models at all.

Some companies have started to provide interpretable ML solutions using proprietary software. While this is a step in the right direction, it is not clear that the proprietary software is better than publicly available software.
For instance, claims made by some companies about performance of their proprietary algorithms are not impressive (e.g., Interpretable AI, whose decision tree performance using mixed integer programming software in 2017 is reported to be often beaten by or comparable to the 1984 Classification and Regression Tree algorithm, CART). 

As discussed earlier, interpretability constraints (like sparsity) lead to optimization problems that have been proven to be computationally hard in the worst case. 
The theoretical hardness of these problems does not mean we cannot solve them, though in real cases, these optimization problems are often difficult to solve. Major improvements have been made in the last decade, and some are discussed later in the Challenges section. Explanation methods, on the other hand, are usually based on derivatives, which lead to easier gradient-based optimization. \\


\noindent\textbf{(iii) Black box models seem to uncover ``hidden patterns.''}

The fact that many scientists have difficulty constructing interpretable models may be fueling the belief that black boxes have the ability to uncover subtle hidden patterns in the data that the user was not previously aware of. A transparent model may be able to uncover these same patterns. If the pattern in the data was important enough that a black box model could leverage it to obtain better predictions, an interpretable model might also locate the same pattern and use it. Again, this depends on the machine learning researcher's ability to create accurate-yet-interpretable models. The researcher needs to create a model that has the capability of uncovering the types of patterns that the user would find interpretable, but also the model needs to be flexible enough to fit the data accurately. This, and the optimization challenges discussed above, are where the difficulty lies with constructing interpretable models.



\section{Encouraging Responsible ML Governance}

Currently the European Union's revolutionary General Data Protection Regulation and other AI regulation plans govern  ``right to an explanation,'' where only an explanation is required, not an interpretable model \citep{goodman2016eu}, in particular ``The data subject shall have the right not to be subject to a decision based solely on automated processing, including profiling, which produces legal effects concerning him or her or similarly significantly affects him or her'' (Article 22 of GDPR regulations from \url{http://www.privacy-regulation.eu/en/22.htm}). If one were to provide an explanation for an automated decision, it is not clear whether the explanation is required to be accurate, complete, or faithful to the underlying model \citep[e.g., see][]{WachterEtAl2018}. Less-than-satisfactory explanations can easily undermine these new policies.

Let us consider a possible mandate that, for certain high-stakes decisions, \textit{no black box should be deployed when there exists an interpretable model with the same level of performance}. If such a mandate were deployed, organizations that produce and sell black box models could then be held accountable if an equally accurate transparent model exists. It could be considered a form of false advertising to sell a black box model if there is an equally-accurate interpretable model. The onus would then fall on organizations to produce black box models only when no transparent model exists for the same task. 

This possible mandate could produce a change in the business model for machine learning. Opacity is viewed as essential in protecting intellectual property, but it is at odds with the requirements of many domains that involve public health or welfare. However, the combination of opacity and explainability is not the only way to incentivize machine learning experts to invest in creating such systems. Compensation for developing an interpretable model could be provided in a lump sum, and the model could be released to the public.
The creator of the model would not be able to profit from licensing the model over a period of time, but the fact that the models are useful for public good applications would make these problems appeal to academics and charitable foundations.  

This proposal will not solve all problems, but it could at least rule out companies selling recidivism prediction models, possibly credit scoring models, and other kinds of models where we can construct accurate-yet-interpretable alternatives. If applied too broadly, it could reduce industrial participation in cases where machine learning might benefit society.

Consider a second proposal, which is weaker than the one provided above, but which might have a similar effect. Let us consider the possibility that organizations that introduce black box models would be mandated to report the accuracy of interpretable modeling methods. In that case, one could more easily determine whether the accuracy/interpretability trade-off claimed by the organization is worthwhile. This also forces the organization to try using interpretable modeling methods. It also encourages the organization to use these methods carefully, otherwise risking the possibility of criticism. 


As mentioned earlier, I have not yet found a high-stakes application where a fully black box model is necessary, despite having worked on many applications. As long as we continue to allow for a broad definition of interpretability that is adapted to the domain, we should be able to improve decision making for serious tasks of societal importance. However, in order for people to design interpretable models, the technology must exist to do so. As discussed earlier, there is a formidable computational hurdle in designing interpretable models, even for standard structured data with already-meaningful features.

\section{Algorithmic Challenges in Interpretable ML}

What if every black box machine learning model could be replaced with one that was equally accurate but also interpretable? If we could do this, we would identify flaws in our models and data that we could not see before. Perhaps we could prevent some of the poor decisions in criminal justice and medicine that are caused by problems with using black box models. We could also eliminate the need for explanations that are misleading and often wrong.

Since interpretability is domain-specific, a large toolbox of possible techniques can come in handy. Below we expand on three of the challenges for interpretable machine learning that appear often. All three cases have something in common: people have been providing interpretable predictive models for these problems for decades, and the human-designed models look just like the type of model we want to create with machine learning. I also discuss some of our current work on these well-known problems.

Each of these challenges is a representative from a major class of models: modeling that uses logical conditions (Challenge 1), linear modeling (Challenge 2), and case-based reasoning (Challenge 3).
 By no means is this set of challenges close to encompassing the large number of domain-specific challenges that exist in creating interpretable models.   

\subsubsection*{Challenge \#1: Constructing optimal logical models}

A logical model consists of statements involving ``or,'' ``and,'' ``if-then,'' etc. The CORELS model in Figure \ref{fig:CORELSModel} is a logical model, called a \textit{rule list}. Decision trees are logical models, as well as conjunctions of disjunctions (``or's'' of ``and's'' -- for instance, IF condition A is true OR conditions B AND C are true, THEN predict yes, otherwise predict no). 

Logical models have been crafted by hand as \textit{expert systems} as far back as the 1970's. Since then, there have been many heuristics for creating logical models; for instance, one might add logical conditions one by one (greedily), and then prune conditions away that are not helpful (again, greedily). These heuristic methods tend to be inaccurate and/or uninterpretable because they do not choose a globally best choice (or approximately best choice) for the logical conditions, and are not designed to be optimally sparse. They might use 200 logical conditions when the same accuracy could be obtained with 5 logical conditions. \citep[C4.5 and CART][decision trees suffer from these problems, as well as a vast number of models from the associative classification literature]{quinlan1993c4,breiman1984classification}. An issue with algorithms that do not aim for optimal (or near-optimal) solutions to optimization problems is that it becomes difficult to tell whether poor performance is due to the choice of algorithm or the combination of the choice of model class and constraints. (Did the algorithm perform poorly because it did not optimize its objective, or because we chose constraints that do not allow enough flexibility in the model to fit the data well?) The question of computing optimal logical models has existed since at least the mid 1990's \citep{AuerHoMa95}. 

We would like models that look like they are created by hand, but they need to be accurate, full-blown machine learning models. To this end, let us consider the following optimization problem, which asks us to find a model that minimizes a combination of the fraction of misclassified training points and the size of the model. Training observations are indexed from $i=1,..,n$, and $\mathcal{F}$ is a family of logical models such as decision trees. The optimization problem is:
\begin{equation}\label{eq:optim}
\min_{f \in \mathcal{F}} \left(\frac{1}{n}\sum_{i=1}^n 1_{[\textrm{training observation}\; i \textrm{ is misclassified by } f]} + \lambda \times \textrm{size}(f)\right).
\end{equation}
Here, the size of the model can be measured by the number of logical conditions in the model, such as the number of leaves in a decision tree. The parameter $\lambda$ is the classification error one would sacrifice in order to have one fewer term in the model; if $\lambda$ is 0.01, it means we would sacrifice 1\% training accuracy in order to reduce the size of the model by one. Another way to say this is that the model would contain an additional term only if this additional term reduced the error by at least 1\%.

The optimization problem in (\ref{eq:optim}) is generally known to be computationally hard. Versions of this optimization problem are some of the fundamental problems of artificial intelligence. The challenge is whether we can solve (or approximately solve) problems like this in practical ways, by leveraging new theoretical techniques and advances in hardware.
 

The model in Figure \ref{fig:CORELSModel} is a machine learning model that comes from the CORELS algorithm \citep{angelino2018}. CORELS solves a special case of (\ref{eq:optim}), for the special choice of $\mathcal{F}$ as the set of rule lists, and where the size of the model is measured by the number of rules in the list. Figure \ref{fig:CORELSModel} has three ``if-then'' rules so its size is 3. In order to minimize (\ref{eq:optim}), CORELS needs to avoid enumerating all possible models, because this would take an extremely long time (perhaps until the end of the universe on a modern laptop for a fairly small dataset).
 The technology underlying the CORELS algorithm was able to solve the optimization problem to optimality in under a minute for the Broward County, FL, dataset discussed above. CORELS' backbone is: (i) a set of theorems allowing massive reductions in the search space of rule lists, (ii) a custom fast bit-vector library that allows fast exploration of the search space, so that CORELS does not need to enumerate all rule lists, and (iii) specialized data structures that keep track of intermediate computations and symmetries. This set of ingredients proved to be a powerful cocktail for handling these tough computational problems.

The example of CORELS enforces two points discussed above, which are, first, that interpretable models sometimes entail hard computational problems, and second, that these computational problems can be solved by leveraging a combination of theoretical and systems-level techniques. CORELS creates one type of logical model; however, there are many more. Formally, \textit{the first challenge is to create algorithms that solve logical modeling problems in a reasonable amount of time, for practical datasets.} 
 
 We have been extending CORELS to more complex problems, such as Falling Rule Lists \citep{wang2015falling,ChenRu2018}, and optimal binary-split decision trees, but there is much work to be done on other types of logical models, with various kinds of constraints.
 
 Note that it is possible to construct interpretable logical models for which the global model is large, and yet each explanation is small. This is discussed in Appendix \ref{appendix:smallerthan}.

\subsubsection*{Challenge \#2: Construct optimal sparse scoring systems}

\begin{figure}[h]
\centering
\includegraphics{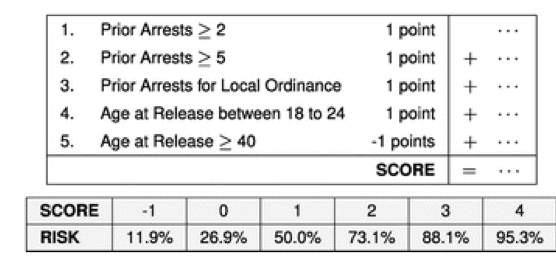}
\caption{Scoring system for risk of recidivism from \citep{RudinUs18} \citep[which grew out of][]{UstunRu2017KDD,ZengUsRu2017,ustun2015slim}. This model was not created by a human; the selection of numbers and features come from the RiskSLIM machine learning algorithm. \label{fig:scoring}}	
\end{figure}

Scoring systems have been designed by hand since at least the Burgess criminological model of 1928 \citep{burgess1928factors}. The Burgess model was designed to predict whether a criminal would violate bail, where individuals received points for being a ``ne'er do well'' or a ``recent immigrant'' that increased their predicted probability of parole violation. (Of course, this model was not created using machine learning, which had not been invented yet.) A scoring system is a sparse linear model with integer coefficients -- the coefficients are the point scores. An example of a scoring system for criminal recidivism is shown in Figure \ref{fig:scoring}, which predicts whether someone will be arrested within 3 years of release. Scoring systems are used pervasively throughout medicine; there are hundreds of scoring systems developed by physicians. Again, the challenge is whether scoring systems -- which look like they could have been produced by a human in the absence of data -- can be produced by a machine learning algorithm, and be as accurate as any other model from any other machine learning algorithm.

There are several ways to formulate the problem of producing a scoring system \citep[see, e.g., ][]{carrizosa2010binarized,SokolovskaEtAl18}. For instance, we could use a special case of (\ref{eq:optim}), where the model size is the number of terms in the model. (Figure \ref{fig:scoring} is a machine learning model with 5 terms.) Sometimes, one can round the coefficients of a logistic regression model to produce a scoring system, but that method does not tend to give accurate models, and does not tend to produce models that have particularly nice coefficients (such as 1 and -1 used in Figure \ref{fig:scoring}). However, solving (\ref{eq:optim}) or its variants is computationally hard, because the domain over which we solve the optimization problem is the integer lattice. (To see this, consider an axis for each of $\{b_1,b_2,...,b_p\}$, where each $b_j$ can take on integer values. This is a lattice that defines the feasible region of the optimization problem.)

The model in Figure \ref{fig:scoring} arose from the solution to a very hard optimization problem. Let us discuss this optimization problem briefly. The goal  is to find the coefficients $b_j$, $j=1...p$ for the linear predictive model $f(\mathbf{z})=\sum_j b_j z_{j}$ where $z_{j}$ is the $j$th covariate of a test observation $\mathbf{z}$. In Figure \ref{fig:scoring}, the $b_j$'s are the point scores, which turned out to be 1, -1, and 0 as a result of optimization, where only the nonzero coefficients are displayed in the figure. In particular, we want to solve:
\[
\min_{b_1,b_2,..,b_p\in \{-10,-9,...,9,10\}} \frac{1}{n}\sum_{i=1}^n \log \left(1+\exp\left(-\sum_{j=1}^p b_j x_{i,j}\right)\right) + \lambda \sum_j 1_{[b_j\neq 0]},
\]
where the point scores $b_j$ are constrained to be integers between -10 and 10, the training observations are indexed by $i=1,...,n$, and $p$ is the total number of covariates for our data. Here the model size is the number of non-zero coefficients, and again $\lambda$ is the trade-off parameter. The first term is the logistic loss used in logistic regression. The problem is hard, specifically it is a mixed-integer-nonlinear program (MINLP) whose domain is the integer lattice.

Despite the hardness of this problem, new cutting plane algorithms have been able to solve this problem to optimality (or near-optimality) for arbitrarily large sample sizes and a moderate number of variables within a few minutes. The latest attempt at solving this problem is the RiskSLIM (Risk-Supersparse-Linear-Integer-Models) algorithm, which is a specialized cutting plane method that adds cutting planes only whenever the solution to a linear program is integer-valued, and otherwise performs branching \citep{UstunRu2017KDD}. 

This optimization problem is similar to what physicians attempt to solve manually, but without writing the optimization problem down like we did above. Because physicians do not use optimization tools to do this, accurate scoring systems tend to be difficult for physicians to create themselves from data. One of our collaborators spent months trying to construct a scoring system himself by adding and removing variables, rounding, and using other heuristics to decide which variables to add, remove, and round. RiskSLIM was useful for helping him with this task \citep{ustun2016adhd}. Formally, 
\textit{the second challenge is to create algorithms for scoring systems that are computationally efficient.} Ideally we would increase the size of the optimal scoring system problems that current methods can practically handle by an order of magnitude. 



\subsubsection*{Challenge \#3 Define interpretability for specific domains and create methods accordingly, including computer vision}

Since interpretability needs to be defined in a domain-specific way, some of the most important technical challenges for the future are tied to specific important domains. Let us start with computer vision, for classification of images. There is a vast and growing body of research on posthoc explainability of deep neural networks, but not as much work in designing \textit{interpretable neural networks}. My goal in this section is to demonstrate that even for classic domains of machine learning, where latent representations of data need to be constructed, there could exist interpretable models that are as accurate as black box models.

For computer vision in particular, there is not a clear definition of interpretability, and the sparsity-related models discussed above do not apply -- sparsity in pixel space does not make sense. There can be many different ideas of what constitutes interpretability, even between different computer vision applications. However, if we can define interpretability somehow for our particular application, we can embed this definition into our algorithm.

Let us define what constitutes interpretability by considering \textit{how people explain to each other} the reasoning processes behind complicated visual classification tasks.
As it turns out, for classification of natural images, domain experts often direct our attention to different parts of the image and explain why these parts of the image were important in their reasoning process. 
\begin{figure}
\centering
\includegraphics[width=0.7\textwidth]{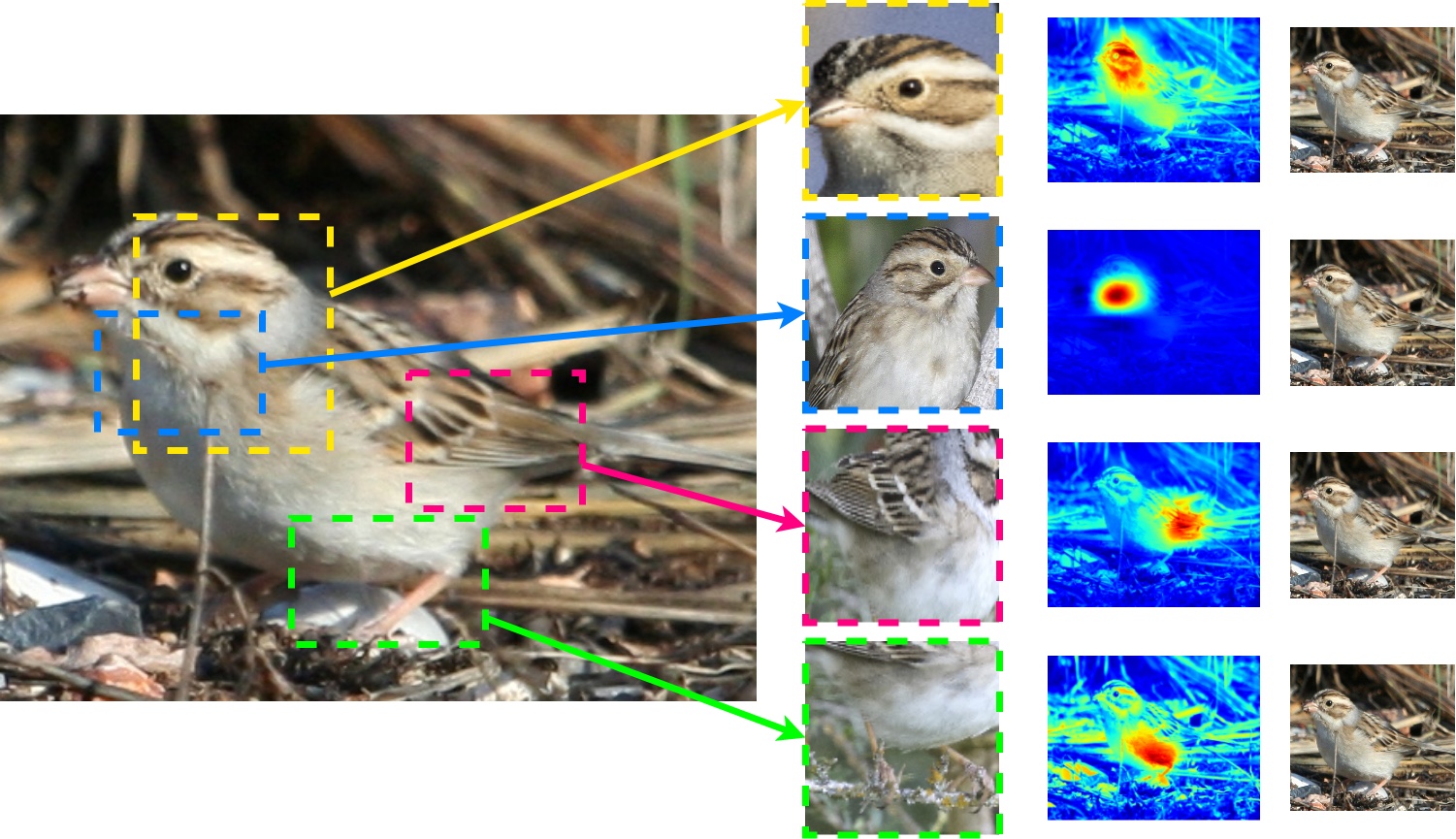}
\caption{Image from the authors of \citep{ChenEtAl18}, indicating that parts of the test image on the left are similar to prototypical parts of training examples. The test image to be classified is on the left, the most similar prototypes are in the middle column, and the heatmaps that show which part of the test image is similar to the prototype are on the right. We included copies of the test image on the right so that it is easier to see what part of the bird the heatmaps are referring to. The similarities of the prototypes to the test image are what determine the predicted class label of the image. Here, the image is predicted to be a clay-colored sparrow. The top prototype seems to be comparing the bird's head to a prototypical head of a clay-colored sparrow, the second prototype considers the throat of the bird, the third looks at feathers, and the last seems to consider the abdomen and leg. Test image from \cite{Omalley}. Prototypes from \cite{ksblack99,Schmierer17,Schmierer15,Schmierer15a}. Image constructed by Alina Barnett. \label{fig:bird}}	
\end{figure}
The question is whether we can construct network architectures for deep learning that can also do this. The network must then make decisions by reasoning about parts of the image so that the explanations are real, and not posthoc.


In a recent attempt to do this, Chen, Li, and colleagues have been building architectures that append a special prototype layer to the end of the network \citep{LiEtAl18,ChenEtAl18}. During training, the prototype layer finds parts of training images that act as prototypes for each class. For instance, for bird classification, the prototype layer might pick out a prototypical head of a blue jay, prototypical feathers of a blue jay, etc. The network also learns a similarity metric between parts of images. Thus, during testing, when a new test image needs to be evaluated, the network finds parts of the test image that are similar to the prototypes it learned during training, as shown in Figure \ref{fig:bird}. The final class prediction of the network is based on the weighted sum of similarities to the prototypes; this is the sum of evidence throughout the image for a particular class. The explanations given by the network are the prototypes (and the weighted similarities to them). These explanations are the actual computations of the model, and these are not posthoc explanations. The network is called ``This look like that'' because its reasoning process considers whether ``this'' part of the image looks like ``that'' prototype.

Training this prototype network is not as easy as training an ordinary neural network; the tricks that have been developed for regular deep learning have not yet been developed for the prototype network. However, so far these prototype networks have been trained to be approximately as accurate as the original black box deep neural networks they were derived from, before the prototype layer was added.

\subsubsection*{Discussion on Interpretability for Specific Domains}

Let us finish this short discussion on challenges to interpretability for specific domains by mentioning that there are vast numbers of papers that have imbued interpretability in their methodology. Interpretability is not mentioned in the title of these papers, and often not in the body of the text. This is why it is almost impossible to create a review article on interpretability in machine learning or statistics without missing the overwhelming majority of it.

It is not clear why review articles for interpretability and explainability make sense to create. We do not normally have reviews of performance/accuracy measures, despite the fact that there are many of them -- accuracy, area under the ROC curve, partial AUC, sensitivity, specificity, discounted cumulative gain, F-score, G-means, and many other domain-specific measures. Interpretability/explainability is just as domain-specific as accuracy performance, so it is not clear why reviews of interpretability make any more sense than reviews of accuracy/performance. I have yet to find even a single recent review that recognized the chasm between interpretability and explainability.

Let us discuss very briefly some of examples of work on interpretability that would not have been covered by recent review articles, and yet are valuable contributions to interpretability in their respective domains. Gallagher et al. \citep{GallagherEtAl17} analyze brain-wide electrical spatiotemporal dynamics to understand depression vulnerability and find interpretable patterns in a low dimensional space. Dimension reduction to interpretable dimensions is an important theme in interpretable machine learning. Problems residing in \textit{applied statistics} are often interpretable because they embed the physics of the domain; e.g., Wang et al. \citep{WangRuMcGo18} create models for recovery curves for prostatectomy patients whose signal and uncertainty obey specific constraints in order to be realistic. Constraints on the uncertainty of the predictions make these models interpretable. 

The setup of the recent 2018 FICO Explainable ML Challenge exemplified the blind belief in the myth of the accuracy/interpretability tradeoff for a specific domain, namely credit scoring. Entrants were instructed to create a black box to predict credit default and explain the model afterwards. \textit{However, there was no performance difference between interpretable models and explainable models for the FICO data.} A globally interpretable model \citep{ChenEtAlFICO2018} won the FICO Recognition Prize for the competition. This is a case where the organizers and judges had not expected an interpretable model to be able to be constructed and thus did not ask entrants to try to construct such a model. The model of \citep{ChenEtAlFICO2018} was an additive model, which is a known form of interpretable model \citep[see also][where additive models are used for medical data]{Lou12, LouCaGeHo13}. Additive models could be optimized using similar techniques to those introduced in Challenge 2 above.

\subsection*{A Technical Reason Why Accurate Interpretable Models Might Exist in Many Domains}
Why is it that accurate interpretable models could possibly exist in so many different domains? Is it really possible that many aspects of nature have simple truths that are waiting to be discovered by machine learning? Although that would be intriguing, I will not make this kind of Occham's-Razor-style argument, in favor of a technical argument about function classes, and in particular, Rashomon Sets. The argument below is fleshed out more formally in \citep{SemenovaPaRu18}. This is related to (but different from) the notation of ``flat minima,'' for which a nice example is given by Hand \citep{Hand}.

Here is the \textit{Rashomon set} argument: Consider that the data permit a large set of reasonably accurate predictive models to exist. Because this set of accurate models is large, it often contains at least one model that is interpretable. This model is thus both interpretable and accurate.

Unpacking this argument slightly, for a given data set, we define the \textit{Rashomon set} as the set of reasonably accurate predictive models (say within a given accuracy from the best model accuracy of boosted decision trees). Because the data are finite, the data could admit many close-to-optimal models that predict differently from each other: a large Rashomon set. I suspect this happens often in practice because sometimes many different machine learning algorithms perform similarly on the same dataset, despite having different functional forms (e.g., random forests, neural networks, support vector machines). As long as the Rashomon set contains a large enough set of models with diverse predictions, it probably contains functions that can be approximated well by simpler functions, and so the Rashomon set can also contain these simpler functions. Said another way, uncertainty arising from the data leads to a Rashomon set, a larger Rashomon set probably contains interpretable models, thus interpretable accurate models often exist.

 If this theory holds, we should expect to see interpretable models exist across domains. These interpretable models may be hard to find through optimization, but at least there is a reason we might expect that such models exist.

If there are many diverse yet good models, it means that algorithms may not be \textit{stable}; an algorithm might choose one model, and a small change to that algorithm or to the dataset may yield a completely different (but still accurate) model. This is not necessarily a bad thing, in fact, the availability of diverse good models means that domain experts may have more flexibility in choosing a model that they find interpretable. Appendix \ref{algstability} discusses this in slightly more detail.

\section{Conclusion}

If this commentary can shift the focus even slightly from the basic assumption underlying most work in Explainable ML -- which is that a black box is necessary for accurate predictions -- we will have considered this document a success. 

If this document can encourage policy makers not to accept black box models without significant attempts at interpretable (rather than explainable) models, that would be even better. 

If we can make people aware of the current challenges right now in interpretable machine learning, it will allow policy-makers the mechanism to demand that more effort should be made in ensuring safety and trust in our machine learning models for high-stakes decisions.

If we do not succeed at these efforts, it is possible that black box models will continue to be permitted when it is not safe to use them. Since the definition of what constitutes a viable explanation is unclear, even strong regulations such as ``right to explanation'' can be undermined with less-than-satisfactory explanations. Further, there will continue to be problems combining black box model predictions with information outside the database, and continued miscalculations of black box model inputs. This may continue to lead to poor decisions throughout our criminal justice system, incorrect safety guidance for air quality disasters, incomprehensible loan decisions, and other widespread societal problems.

\section*{Acknowledgments}
I would like to thank Fulton Wang, Tong Wang, Chaofan Chen, Oscar Li, Alina Barnett, Tom Dietterich, Margo Seltzer, Elaine Angelino, Nicholas Larus-Stone, Elizabeth Mannshart, Maya Gupta, and several others who helped my thought processes in various ways, and particularly Berk Ustun, Ron Parr, Rob Holte, and my father, Stephen Rudin, who went to considerable efforts to provide thoughtful comments and discussion. I would also like to thank two anonymous reviewers for their suggestions that improved the manuscript. I would like to acknowledge funding from the Laura and John Arnold Foundation, NIH, NSF, DARPA, the Lord Foundation of North Carolina, and MIT-Lincoln Laboratory.


\small{
\bibliography{EditInterp}

\begin{thebibliography}{10}

\bibitem{nyt-computers-crim-justice}
Wexler R.
\newblock When a Computer Program Keeps You in Jail: How Computers are Harming
  Criminal Justice.
\newblock New York Times. 2017 June 13;.

\bibitem{McGough2018}
McGough M.
\newblock How bad is Sacramento's air, exactly? Google results appear at odds
  with reality, some say.
\newblock Sacramento Bee. 2018 August 7;.

\bibitem{VarshneyAl2016}
Varshney KR, Alemzadeh H.
\newblock On the safety of machine learning: Cyber-physical systems, decision
  sciences, and data products.
\newblock Big Data. 2016 10;5.

\bibitem{Freitas:2014ic}
Freitas AA.
\newblock {Comprehensible classification models: a position paper}.
\newblock ACM SIGKDD Explorations Newsletter. 2014 Mar;15(1):1--10.

\bibitem{kodratoff1994comprehensibility}
Kodratoff Y.
\newblock The comprehensibility manifesto.
\newblock KDD Nugget Newsletter. 1994;94(9).

\bibitem{huysmans:2011gq}
Huysmans J, Dejaeger K, Mues C, Vanthienen J, Baesens B.
\newblock An empirical evaluation of the comprehensibility of decision table,
  tree and rule based predictive models.
\newblock Decision Support Systems. 2011;51(1):141--154.

\bibitem{ruping2006learning}
R{\"u}ping S.
\newblock Learning Interpretable Models.
\newblock Universit{\"a}t Dortmund; 2006.

\bibitem{gupta2016monotonic}
Gupta M, Cotter A, Pfeifer J, Voevodski K, Canini K, Mangylov A, et~al.
\newblock Monotonic calibrated interpolated look-up tables.
\newblock Journal of Machine Learning Research. 2016;17(109):1--47.

\bibitem{LouCaGeHo13}
Lou Y, Caruana R, Gehrke J, Hooker G.
\newblock Accurate Intelligible Models with Pairwise Interactions.
\newblock In: Proceedings of 19th ACM SIGKDD International Conference on
  Knowledge Discovery and Data Mining ({KDD}). ACM; 2013. .

\bibitem{miller1956magical}
Miller G.
\newblock The magical number seven, plus or minus two: Some limits on our
  capacity for processing information.
\newblock The Psychological Review. 1956;63:81--97.

\bibitem{cowan2010magical}
Cowan N.
\newblock The magical mystery four how is working memory capacity limited, and
  why?
\newblock Current directions in psychological science. 2010;19(1):51--57.

\bibitem{Wiens2018}
Wang J, Oh J, Wang H, Wiens J.
\newblock Learning Credible Models.
\newblock In: Proceedings of 24th ACM SIGKDD International Conference on
  Knowledge Discovery and Data Mining ({KDD}). ACM; 2018. p. 2417--2426.

\bibitem{Rudin18}
Rudin C.
\newblock Please Stop Explaining Black Box Models for High Stakes Decisions.
\newblock In: Proceedings of {NeurIPS} 2018 Workshop on Critiquing and
  Correcting Trends in Machine Learning; 2018. .

\bibitem{Holte93}
Holte RC.
\newblock Very simple classification rules perform well on most commonly used
  datasets.
\newblock Machine Learning. 1993;11(1):63--91.

\bibitem{Fayyad96fromdata}
Fayyad U, Piatetsky-Shapiro G, Smyth P.
\newblock From data mining to knowledge discovery in databases.
\newblock AI Magazine. 1996;17:37--54.

\bibitem{crispdm}
Chapman P, et~al.
\newblock {CRISP-DM} 1.0 - Step-by-step data mining guide.
\newblock SPSS; 2000.

\bibitem{bigdata}
Agrawal D, Bernstein P, Bertino E, Davidson S, Dayal U, Franklin M, et~al.
\newblock Challenges and Opportunities with Big Data: A white paper prepared
  for the Computing Community Consortium committee of the Computing Research
  Association; 2012.
\newblock Available from:
  \url{http://cra.org/ccc/resources/ccc-led-whitepapers/}.

\bibitem{XAIBAA}
{Defense Advanced Research Projects Agency}. Broad Agency Announcement,
  Explainable Artificial Intelligence (XAI), {DARPA-BAA}-16-53; 2016.
\newblock Published August 10.
\newblock Available from
  \url{https://www.darpa.mil/attachments/DARPA-BAA-16-53.pdf}.

\bibitem{Hand}
Hand D.
\newblock Classifier Technology and the Illusion of Progress.
\newblock Statist Sci. 2006;21(1):1--14.

\bibitem{RudinETAL2010}
Rudin C, Passonneau R, Radeva A, Dutta H, Ierome S, Isaac D.
\newblock A Process for Predicting Manhole Events In {M}anhattan.
\newblock {M}achine {L}earning. 2010;80:1--31.

\bibitem{RudinUs18}
Rudin C, Ustun B.
\newblock Optimized Scoring Systems: Toward Trust in Machine Learning for
  Healthcare and Criminal Justice.
\newblock Interfaces. 2018;48:399--486.
\newblock Special Issue: 2017 Daniel H. Wagner Prize for Excellence in
  Operations Research Practice September-October 2018.

\bibitem{ChenEtAlFICO2018}
Chen C, Lin K, Rudin C, Shaposhnik Y, Wang S, Wang T.
\newblock An Interpretable Model with Globally Consistent Explanations for
  Credit Risk.
\newblock In: Proceedings of NeurIPS 2018 Workshop on Challenges and
  Opportunities for AI in Financial Services: the Impact of Fairness,
  Explainability, Accuracy, and Privacy; 2018. .

\bibitem{MittelstadtEtAl19}
Mittelstadt B, Russell C, Wachter S.
\newblock Explaining Explanations in AI.
\newblock In: In Proceedings of Fairness, Accountability, and Transparency
  ({FAT*}); 2019. .

\bibitem{Flores16}
Flores AW, Lowenkamp CT, Bechtel K.
\newblock False Positives, False Negatives, and False Analyses: A Rejoinder to
  ``{M}achine Bias: There's Software Used Across the Country to Predict Future
  Criminals''.
\newblock Federal probation. 2016 September;80(2):38--46.

\bibitem{propublica2016}
Angwin J, Larson J, Mattu S, Kirchner L. Machine Bias.
\newblock ProPublica; 2016.
\newblock Available from:
  \url{https://www.propublica.org/article/machine-bias-risk-assessments-in-criminal-sentencing}.

\bibitem{LarsonMaKiAn16}
Larson J, Mattu S, Kirchner L, Angwin J. How We Analyzed the {COMPAS}
  Recidivism Algorithm.
\newblock ProPublica; 2016.
\newblock
  \url{https://www.propublica.org/article/how-we-analyzed-the-compas-recidivism-algorithm}.

\bibitem{RudinWaCo18}
Rudin C, Wang C, Coker B.
\newblock {The age of secrecy and unfairness in recidivism prediction}.
\newblock arXiv e-prints 1811 00731 $[$applied statistics$]$. 2018 Nov;.

\bibitem{Checkermallow}
Checkermallow. Canis lupus winstonii (Siberian Husky); 2016.
\newblock Public domain image.
\newblock
  \url{https://www.flickr.com/photos/132792051@N06/28302196071/in/photolist-K7Y9RM-utZTV9-QWJmHo-QAEdSE-QAE3pL-TvjNJu-
  tziyrj-EWFwEx-DWb7T4-DTRAWu-CYLBpP-DMUVn2-dUbgLG-ccuabw-57nNvJ-UpDv4D-eNyCQP-q8aWpJ-86gced-QLBwiG-QP7k6v-aNxiRc-rmTdLW-oeTM8i-d1rkCG-ueSwz4-
  dYKwJx-7PxAPF-KFUqKN-TkarEj-7X5FZ2-7WS6Z2-7X5Gwa-7X5GkT-7Z8w5s-s4St8A-
  qsa12b-7X8Vqs-7X8VLy-7X5Gm6-7X5Gjp-PTy69W-7X8VQ3-7X8VEy-7X5GqD-iaMjUN-7X8VgE-odbiWy-TkacgQ-7X5Gk4/}.

\bibitem{northpointe}
Brennan T, Dieterich W, Ehret B.
\newblock Evaluating the Predictive Validity of the {COMPAS} Risk and Needs
  Assessment System.
\newblock Criminal Justice and Behavior. 2009 January;36(1):21--40.

\bibitem{ZengUsRu2017}
Zeng J, Ustun B, Rudin C.
\newblock Interpretable classification models for recidivism prediction.
\newblock Journal of the Royal Statistical Society: Series {A} (Statistics in
  Society). 2017;180(3):689--722.

\bibitem{tollenaar2013method}
Tollenaar N, {van der Heijden} PGM.
\newblock Which method predicts recidivism best?: a comparison of statistical,
  machine learning and data mining predictive models.
\newblock Journal of the Royal Statistical Society: Series A (Statistics in
  Society). 2013;176(2):565--584.

\bibitem{angelino2018}
Angelino E, Larus-Stone N, Alabi D, Seltzer M, Rudin C.
\newblock Certifiably optimal rule lists for categorical data.
\newblock Journal of Machine Learning Research. 2018;19:1--79.

\bibitem{Mannshardt2018}
Mannshardt E, Naess L.
\newblock Air quality in the {USA}.
\newblock Significance. 2018 Oct;15:24--27.

\bibitem{Zech2018}
Zech JR, et~al.
\newblock Variable generalization performance of a deep learning model to
  detect pneumonia in chest radiographs: A cross-sectional study.
\newblock PLoS Med. 2018;15(e1002683).

\bibitem{ChangEtAl2012}
Chang A, Rudin C, Cavaretta M, Thomas R, Chou G.
\newblock How to Reverse-Engineer Quality Rankings.
\newblock Machine Learning. 2012 September;88:369--398.

\bibitem{goodman2016eu}
Goodman B, Flaxman S.
\newblock {EU} regulations on algorithmic decision-making and a `right to
  explanation'.
\newblock AI Magazine. 2017;38(3).

\bibitem{WachterEtAl2018}
Wachter S, Mittelstadt B, Russell C.
\newblock Counterfactual Explanations without Opening the Black Box: Automated
  Decisions and the {GDPR}.
\newblock Harvard Journal of Law \& Technology. 2018;1(2).

\bibitem{quinlan1993c4}
Quinlan JR.
\newblock C4. 5: programs for machine learning. vol.~1.
\newblock Morgan Kaufmann; 1993.

\bibitem{breiman1984classification}
Breiman L, Friedman J, Stone CJ, Olshen RA.
\newblock Classification and regression trees.
\newblock CRC press; 1984.

\bibitem{AuerHoMa95}
Auer P, Holte RC, Maass W.
\newblock Theory and Applications of Agnostic PAC-Learning with Small Decision
  Trees.
\newblock In: Machine Learning Proceedings 1995. San Francisco (CA): Morgan
  Kaufmann; 1995. p. 21 -- 29.

\bibitem{wang2015falling}
Wang F, Rudin C.
\newblock Falling Rule Lists.
\newblock In: Proceedings of Machine Learning Research Vol. 38: Artificial
  Intelligence and Statistics ({AISTATS}); 2015. p. 1013--1022.

\bibitem{ChenRu2018}
Chen C, Rudin C.
\newblock An optimization approach to learning falling rule lists.
\newblock In: Proceedings of Machine Learning Research Vol. 84: Artificial
  Intelligence and Statistics {(AISTATS)}; 2018. p. 604--612.

\bibitem{burgess1928factors}
Burgess EW. Factors determining success or failure on parole; 1928.
\newblock Illinois Committee on Indeterminate-Sentence Law and Parole
  Springfield, IL.

\bibitem{UstunRu2017KDD}
Ustun B, Rudin C.
\newblock Optimized Risk Scores.
\newblock In: Proceedings of the 23rd {ACM} {SIGKDD} International Conference
  on Knowledge Discovery and Data Mining ({KDD}); 2017. .

\bibitem{ustun2015slim}
Ustun B, Rudin C.
\newblock Supersparse linear integer models for optimized medical scoring
  systems.
\newblock Machine Learning. 2015;p. 1--43.

\bibitem{carrizosa2010binarized}
Carrizosa E, Mart{\'\i}n-Barrag{\'a}n B, Morales DR.
\newblock Binarized support vector machines.
\newblock INFORMS Journal on Computing. 2010;22(1):154--167.

\bibitem{SokolovskaEtAl18}
Sokolovska N, Chevaleyre Y, Zucker JD.
\newblock A Provable Algorithm for Learning Interpretable Scoring Systems.
\newblock In: Proceedings of Machine Learning Research Vol. 84: Artificial
  Intelligence and Statistics ({AISTATS}); 2018. p. 566--574.

\bibitem{ustun2016adhd}
Ustun B, et~al.
\newblock {The World Health Organization Adult Attention-Deficit/Hyperactivity
  Disorder Self-Report Screening Scale for DSM-5}.
\newblock JAMA Psychiatry. 2017;74(5):520--526.

\bibitem{ChenEtAl18}
Chen C, Li O, Tao C, Barnett A, Su J, Rudin C.
\newblock \textit{This} Looks Like \textit{that}: Deep Learning for
  Interpretable Image Recognition.
\newblock In: Neural Information Processing Systems {(NeurIPS)}; 2019. .

\bibitem{Omalley}
O'Malley D. Clay-colored Sparrow; 2014.
\newblock Public domain image.
\newblock \url{https://www.flickr.com/photos/62798180@N03/11895857625/}.

\bibitem{ksblack99}
ksblack99. Clay-colored Sparrow; 2018.
\newblock Public domain image.
\newblock \url{https://www.flickr.com/photos/ksblack99/42047311831/}.

\bibitem{Schmierer17}
Schmierer A. Clay-colored Sparrow; 2017.
\newblock Public domain image.
\newblock \url{https://flic.kr/p/T6QVkY}.

\bibitem{Schmierer15}
Schmierer A. Clay-colored Sparrow; 2015.
\newblock Public domain image.
\newblock \url{https://flic.kr/p/rguC7K}.

\bibitem{Schmierer15a}
Schmierer A. Clay-colored Sparrow; 2015.
\newblock Public domain image.
\newblock \url{https://www.flickr.com/photos/sloalan/16585472235/ }.

\bibitem{LiEtAl18}
Li O, Liu H, Chen C, Rudin C.
\newblock Deep Learning for Case-based Reasoning through Prototypes: A Neural
  Network that Explains its Predictions.
\newblock In: Proceedings of {AAAI} Conference on Artificial Intelligence
  {({AAAI})}; 2018. p. 3530--3537.

\bibitem{GallagherEtAl17}
Gallagher N, et~al.
\newblock Cross-Spectral Factor Analysis.
\newblock In: Proceedings of Advances in Neural Information Processing Systems
  30 {(NeurIPS)}. Curran Associates, Inc.; 2017. p. 6842--6852.

\bibitem{WangRuMcGo18}
Wang F, Rudin C, Mccormick TH, Gore JL.
\newblock Modeling recovery curves with application to prostatectomy.
\newblock Biostatistics. 2018;p. kxy002.
\newblock Available from: \url{http://dx.doi.org/10.1093/biostatistics/kxy002}.

\bibitem{Lou12}
Lou Y, Caruana R, Gehrke J.
\newblock Intelligible Models for Classification and Regression.
\newblock In: Proceedings of Knowledge Discovery in Databases ({KDD}). ACM;
  2012. .

\bibitem{SemenovaPaRu18}
Semenova L, Parr R, Rudin C.
\newblock A study in {R}ashomon curves and volumes: A new perspective on
  generalization and model simplicity in machine learning; 2018.
\newblock In progress.

\bibitem{Razavian15}
Razavian N, et~al.
\newblock Population-Level Prediction of Type 2 Diabetes From Claims Data and
  Analysis of Risk Factors.
\newblock Big Data. 2015;3(4).

\bibitem{UstunSpLi18}
{Ustun} B, {Spangher} A, {Liu} Y.
\newblock {Actionable Recourse in Linear Classification}.
\newblock In: {ACM} Conference on Fairness, Accountability and Transparency
  ({FAT*}); 2019. .

\bibitem{SuEtAl16}
Su G, Wei D, Varshney KR, Malioutov DM.
\newblock Interpretable Two-Level Boolean Rule Learning for Classification.
\newblock In: Proceedings of {ICML} Workshop on Human Interpretability in
  Machine Learning; 2016. p. 66--70.

\bibitem{DashEtAl18}
Dash S, G\"{u}nl\"uk O, Wei D.
\newblock Boolean Decision Rules via Column Generation.
\newblock In: 32nd Conference on Neural Information Processing Systems
  ({NeurIPS}); 2018. .

\bibitem{WangEtAl2017}
Wang T, Rudin C, Doshi-Velez F, Liu Y, Klampfl E, MacNeille P.
\newblock A Bayesian Framework for Learning Rule Sets for Interpretable
  Classification.
\newblock Journal of Machine Learning Research. 2017;18(70):1--37.

\bibitem{Rijnbeek10}
Rijnbeek PR, Kors JA.
\newblock Finding a Short and Accurate Decision Rule in Disjunctive Normal Form
  by Exhaustive Search.
\newblock Machine Learning. 2010 Jul;80(1):33--62.

\bibitem{goh2014box}
Goh ST, Rudin C.
\newblock Box Drawings for Learning with Imbalanced Data.
\newblock In: Proceedings of the 20th ACM SIGKDD Conference on Knowledge
  Discovery and Data Mining {(KDD)}; 2014. .

\bibitem{MurdocEtAl19}
{Murdoch} WJ, {Singh} C, {Kumbier} K, {Abbasi-Asl} R, {Yu} B.
\newblock Interpretable machine learning: definitions, methods, and
  applications.
\newblock arXiv e-prints: 1901 04592 $[$statistical machine learning$]$. 2019
  Jan;.

\end{thebibliography}
} 


\newpage
\appendix

\section{On the Two Types of Black Box}\label{twotypes}
Black box models of the first type are too complicated for a human to comprehend, and black box models of the second type are proprietary. Some models are of both types. The consequences of these two types of black box are different, but related. For instance, for a black box model that is complicated but not proprietary, we at least know what variables it uses. We also know the model form and could use that to attempt to analyze the different parts of the model. For a black box model that is proprietary but not complicated \citep[we have evidence that COMPAS is such a model,][]{RudinWaCo18}, we may not even have access to query it in order to study it. If a proprietary model is too sparse, there is a risk that it could be easily reverse-engineered, thus there is an incentive to make proprietary models complicated in order to preserve their secrecy.  


\section{Performance Comparisons}
\label{AppendixSame}
For most problems with meaningful structured covariates, machine learning algorithms tend to perform similarly, with no algorithm clearly dominating the others. The variation due to tuning parameters of a single algorithm can often be higher than the variation between algorithms. 
This lack of single dominating algorithm for structured data is arguably why the field of machine learning focuses on image and speech recognition, whose data are represented in raw features (pixels, sound files); these are fields for which the choice of algorithm impacts performance. Even for complex domains such as medical records, it has been reported in some studies that logistic regression has identical performance to deep neural networks \citep[e.g.][]{Razavian15}. 

If there is no dominating algorithm, the Rashomon Set argument discussed above would suggest that interpretable models might perform well. 

Unfortunately the culture of publication within machine learning favors selective reporting of algorithms on selectively chosen datasets. Papers are often rejected if small or no performance gains are reported between algorithms. This encourages omission of accurate baselines for comparison, as well as omission of datasets on which the method does not perform well, and encourages authors to poorly tune the parameters of baseline methods, or equivalently, place more effort into tuning the parameters of the author's own method. This creates an illusion of large performance differences between algorithms, even when such performance differences do not truly exist.

\section{Counterfactual Explanations}
\label{appendix:counterfactual}

Some have argued that counterfactual explanations \citep[e.g., see][]{WachterEtAl2018} are a way for black boxes to provide useful information while preserving secrecy of the global model. Counterfactual explanations, also called inverse classification, state a change in features that is sufficient (but not necessary) for the prediction to switch to another class (e.g., ``If you reduced your debt by \$5000 and increased your savings by \$50\% then you would have qualified for the loan you applied for''). This is important for recourse in certain types of decisions, meaning that the user could take an action to reverse a decision \citep{UstunSpLi18}.

There are several problems with the argument that counterfactual explanations are sufficient. For loan applications, for instance, we would want the counterfactual explanation to provide the \textit{lowest cost} action for the user to take, \textit{according to the user's own cost metric}. \citep[See][for an example of lowest-cost counterfactual reasoning in product rankings]{ChangEtAl2012}. In other words, let us say that there is more than one counterfactual explanation available (e.g., the first explanation is ``If you reduced your debt by \$5000 and increased your savings by \$50\% then you would have qualified for the loan you applied for'' and the second explanation is ``If you had gotten a job that pays \$500 more per week, then you would have qualified for the loan''). In that case, the explanation shown to the user should be the easiest one for the user to actually accomplish. However, it is unclear in advance which explanation would be easier for the user to accomplish. In the credit example, perhaps it is easier for the user to save money rather than get a job or vice versa. In order to determine which explanation is the lowest cost for the user, we would need to elicit cost information for the user, and that cost information is generally very difficult to obtain; worse, the cost information could actually change as the user attempts to follow the policy provided by the counterfactual explanation (e.g., it turns out to be harder than the user thought to get a salary increase). For that reason it is unclear that counterfactual explanations would suffice for high stakes decisions. Additionally, counterfactual explanations of black boxes have many of the other pitfalls discussed throughout this paper.

\section{Interpretable Models that Provide Smaller-Than-Global Explanations}\label{appendix:smallerthan}

It is possible to create a global model (perhaps a complicated one) for which explanations for any given individual are very sparse. In other words, even if the global model would take several pages of text to write, the prediction for a given individual can be very simple to calculate (perhaps requiring only 1-2 conditions). Let us consider the case of credit risk prediction. Assume we do not need to justify to the client why we would grant a loan, but we would need to justify why we would deny a loan. 

Let us consider a disjunctive normal form model, which is a collection of ``or's'' of ``and's.'' For instance, the model might deny a loan if ``(credit history too short AND at least one bad past trade) OR (at least 4 bad past trades) OR (at least one recent delinquency AND high percentage of delinquent trades).'' Even if we had hundreds of conjunctions within the model, only one of these needs to be shown to the client; if any conjunction is true, that conjunction is a defining reason why the client would be denied a loan. In other words, if the client had ``at least one recent delinquency AND high percentage of delinquent trades,'' then regardless of any other aspects of her credit history, she could be shown that simple explanation, and it would be a defining reason why her loan application would be denied. 

Disjunctive normal form models are well-studied, and are called by various names, such as ``or's of and's,'' as well as ``decision rules,'' ``rule sets'' and ``associative classifiers.'' 
There has been substantial work in being able to generate such models over the past few years so that the models are globally interpretable, not just locally interpretable (meaning that the global model consists of a small number of conjunctions) \citep[e.g., see][]{SuEtAl16,DashEtAl18,WangEtAl2017,Rijnbeek10,goh2014box}.

There are many other types of models that would provide smaller-than-global explanations. For instance, falling rule lists \citep{wang2015falling,ChenRu2018} provide shorter explanations for the decisions that are most important. For instance, a falling rule list for predicting patient mortality would use few logical conditions to categorize whether a patient is in a high-risk group, but use several additional logical conditions to determine which low-risk group a patient falls into.

\section{Algorithm Stability}\label{algstability}
A common criticism of decision trees is that they are not stable, meaning that small changes in the training data lead to completely different trees, giving no guidance as to which tree to choose.
In fact, the same problem can happen in \textit{linear} models when there are highly correlated features. This can happen even in basic least squares, where correlations between features can lead to very different models having precisely the same levels of performance. When there are correlated features, the lack of stability happens with most algorithms that are not strongly regularized.  

  I hypothesize this instability in the learning algorithm could be a side-effect of the Rashomon effect mentioned earlier -- that there are many different almost-equally good predictive models. Adding regularization to an algorithm increases stability, but also limits flexibility of the user to choose which element of the Rashomon set would be more desirable.  
  
  For applications where the models are purely predictive and not causal (e.g., in criminal recidivism where we use age and prior criminal history to predict future crime), there is no assumption that the model represents how outcomes are actually generated. The importance of the variables in the model does not reflect a causal relationship between the variables and the outcomes. Thus, without additional guidance from the domain expert, there is no way to proceed further to choose a single ``best model'' among the set of models that perform similarly. As discussed above, regularization can act as this additional input.

I view the lack of algorithmic stability as an advantage rather than a disadvantage. If the lack of stability is indeed caused by a large Rashomon effect, it means that domain experts can add more constraints to the model to customize it without losing accuracy.

In other words, while many people criticize methods such as decision trees for not being stable, I view that as a strength of interpretability for decision trees. If there are many equally accurate trees, the domain expert can pick the one that is the most interpretable.

Note that not all researchers working in interpretability agree with this general sentiment about the advantages of instability \citep{MurdocEtAl19}.

\end{document}